\documentclass[letterpaper, 10 pt, conference]{ieeeconf}  
\pdfoutput=1 
\IEEEoverridecommandlockouts                              
\usepackage{gensymb}
\usepackage{cclicenses, graphicx}
\usepackage{amsmath}
\usepackage{array}
\newcolumntype{C}[1]{>{\centering\arraybackslash}m{#1}}

\title{\LARGE \bf EMG-Controlled Non-Anthropomorphic Hand Teleoperation \\ Using a  Continuous Teleoperation Subspace}

\author{Cassie Meeker$^{1}$ and Matei Ciocarlie$^{1}$%
\thanks{*
This work was supported in part by the 
ONR Young Investigator Program award N00014-16-1-2026.}%
\thanks{$^{1}$Department of Mechanical Engineering, Columbia University, New York, NY 10027, USA.}%
\thanks{\hspace{-3mm}{\tt\small \{cgm2144, matei.ciocarlie\}@columbia.edu}}%
}

\begin{document}

\maketitle
\thispagestyle{empty}
\pagestyle{empty}

\begin{abstract}
We present a method for EMG-driven teleoperation of non-anthropomorphic robot hands. EMG sensors are appealing as a wearable, inexpensive, and unobtrusive way to  gather information about the teleoperator's hand pose. However, mapping from EMG signals to the pose space of a non-anthropomorphic hand presents multiple challenges. We present a method that first projects from forearm EMG into a subspace relevant to teleoperation. To increase robustness, we use a model which combines continuous and discrete predictors along different dimensions of this subspace. We then project from the teleoperation subspace into the pose space of the robot hand. Our method is effective and intuitive, as it enables novice users to teleoperate pick and place tasks faster and more robustly than state-of-the-art EMG teleoperation methods when applied to a non-anthropomorphic, multi-DOF robot hand.

\end{abstract}

\section{Introduction}

In robotic manipulation, the space of possible scenarios and objects that can be encountered is very large. One solution is to deploy robotic manipulators as part of a human-robot collaborative team. Through teleoperation, a human's cognitive abilities can be exploited to deal with decisions that are particularly difficult for autonomy.

In many cases, particularly in emergency or disaster response scenarios, it is unreasonable to expect that people who need to guide robots through manipulation tasks will be roboticists, or even expert users. Intuitive teleoperation controls are desirable because they allow non-expert users to complete tasks quickly and safely.

Harvesting the natural movement of a human hand to control a robot hand can provide an intuitive control~\cite{ferre2007}. However, this approach is difficult to use with non-anthropomorphic robot hands, which have very different kinematics than their human counterparts. Still, as non-anthropomorphic hands have proven to be versatile, robust, and cost effective, previous research has proposed mappings aiming to resolve these kinematic differences. These mappings include fingertip~\cite{rohling1993}, joint~\cite{cerulo2017}, and pose mapping~\cite{ekvall2004}, as well as our own teleoperation subspace mapping~\cite{meeker2018}. 

All of the above mappings require the pose of the teleoperator's hand as input. Typically this information is obtained through use of instrumented datagloves or vision. While datagloves and vision are robust, vision-based methods often require environments which are well-lit and which do not have many obstacles that occlude the hand. Datagloves can interfere with a hand's fine manipulation or tactile sensing abilities, and are easily damaged. We would like a control input that collects information about the teleoperator's hand pose while being \textit{wearable, inexpensive, and unobtrusive}. 

One such input is forearm electromyography (EMG). EMG is low profile, and its position on the forearm, rather than the hand, makes it less susceptible to damage while leaving the hand completely unencumbered. Inexpensive EMG armbands are quickly becoming commodity products. However, to use EMG as a control for teleoperation, we must find a way to map between forearm EMG signals and the pose of a non-anthropomorphic robot hand.

\begin{figure}[t]
\centering
\vspace{0mm}
\begin{tabular}{r}
\hspace{-4mm}
\includegraphics[trim=3.3cm 6.5cm 2.2cm 4.5cm, clip, width=1.03\linewidth]{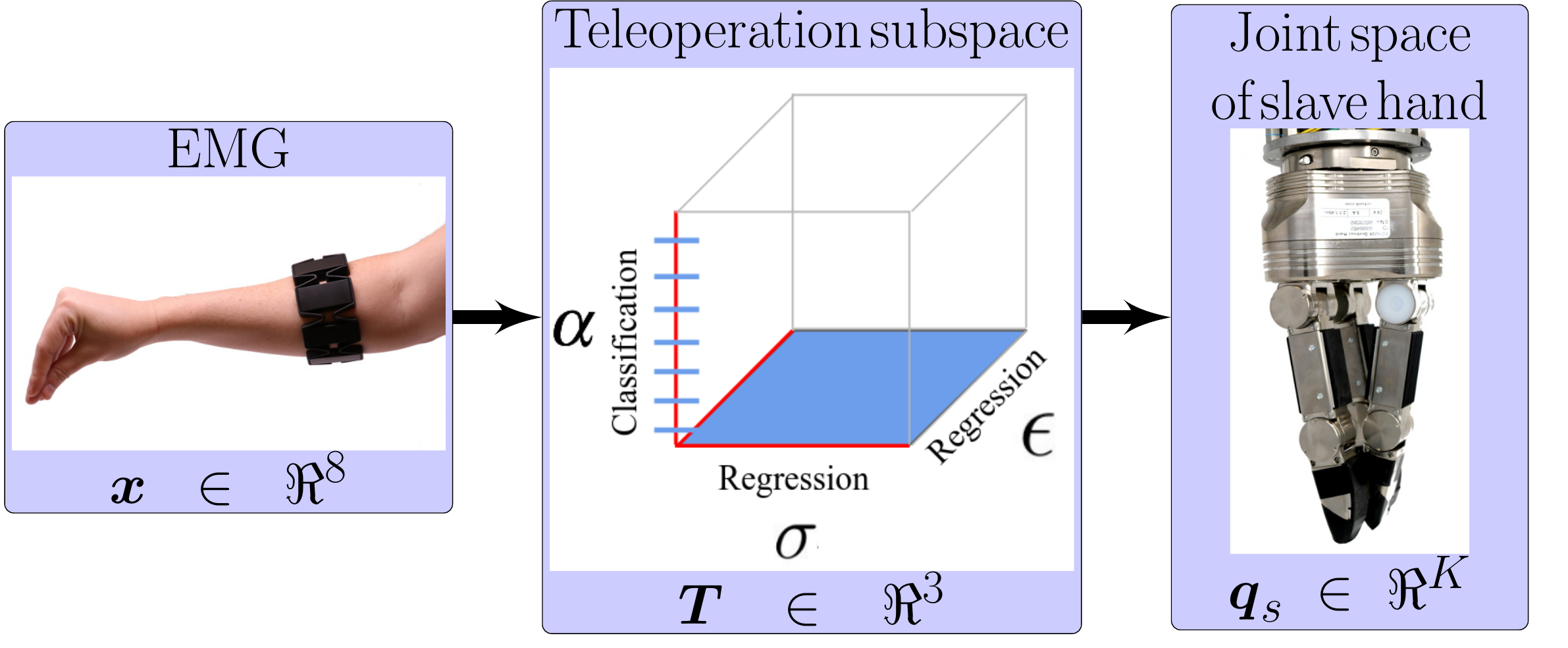}
\end{tabular}
\vspace{-4mm}
\caption{Teleoperation control using EMG and teleoperation subspace}
\label{fig:emg_mapping}
\vspace{-5mm}
\end{figure}

Here we use a low dimensional subspace to teleoperate a non-anthropomorphic hand. We have shown that this subspace is useful for teleoperation in previous work~\cite{meeker2018}; however, our previous formulation required a dataglove for teleoperation control. In this work, we project forearm EMG into this subspace and use it as a control for a fully actuated, multi-degree-of-freedom (DOF) robot (Figure~\ref{fig:emg_mapping}).  

To our knowledge, we are the first to demonstrate teleoperation of a non-anthropomorphic, multi-DOF robot hand using forearm EMG as a control input. The key to our method is our use of a hybrid model, which is a combination of discrete and continuous predictors, along with a continuous pose subspace which we can span and which is independent of training. These elements allow users to control a robot using natural hand movements, making the control intuitive. 

Other state-of-the-art teleoperation methods have used forearm EMG as a control input but have not demonstrated control of non-anthropomorphic, multi-DOF robot hands. We modified several state-of-the-art teleoperation controls as necessary to make them applicable to these kinds of hands and compare them to our method. Experiments with novice teleoperators prove that our method can grasp a wide variety of objects, and do so faster than state-of-the-art methods.

\section{Related Work}\label{sec:related_work}

EMG control in teleoperation has been primarily used for two applications: grippers and control of anthropomorphic hands, principally in the context of prosthetics. 

EMG teleoperation of a one DOF gripper usually uses proportional control to determine the gripper aperture~\cite{herrera2004}\cite{gillespie2010}.  Using EMG, it is possible to estimate the aperture of a gripper in conjunction with the position of the elbow and the wrist~\cite{vasan2017}. Choudhary et al. developed a three DOF hand with EMG control, but the EMG only provides a binary open/close signal~\cite{choudhary2012}. EMG control for grippers is consistent, intuitive and reliable, but grippers do not give the sort of versatility that many complex grasping scenarios require. 

The other context in which EMG teleoperation has been studied is prosthetics. Prosthetic controls can use an agonistic/antagonistic muscle pair to control a single DOF underactuated hand~\cite{fani2016} or they can estimate the motion of one of the joints of the human hand~\cite{malesevic2017}~\cite{smith2008}. Studies which control more than one DOF usually control the wrist, often in a continuous and proportional manner~\cite{hahne2014}\cite{jiang2009}\cite{jiang2014}\cite{lin2017}.

One interesting approach combines force control and gesture control. Yoshikawa et al. classified hand position, and then used an empirical EMG-joint angle model to estimate wrist or metacarpophalangeal (MCP) joint angles~\cite{yoshikawa2007}. Yamanoi et al. developed models of the relationship between force and EMG for multiple postures. They used EMG to classify hand pose, and then determined grip force based on the force model~\cite{yamanoi2017}. Castellini et al. used a similar strategy in a data based approach~\cite{castellini2009}. Gijsberts et al. created an entirely force driven EMG control for a prosthetic hand~\cite{gijsberts2014}.

To control multi-DOF anthropomorphic robots, it is common to use low dimensional spaces and dimension reduction~\cite{santello2016}. 
Rossi et al. mapped EMG signals to the synergies of an underactuated robot hand~\cite{rossi2017}. Matrone et al. used agonistic/antagonistic muscle pairs to determine wrist position, and then control a robotic hand using the synergies of the robot~\cite{matrone2012}.  Artemiadis and Kyriakopoulos used principal component analysis (PCA) to find low-dimensional representations of both kinematic and EMG data for the human arm.  They mapped between the two spaces to teleoperate a robot arm~\cite{artemiadis2011}, and, later, a hand-arm system~\cite{liarokapis2013}. However, fully anthropomorphic robot hands are complex, fragile and expensive. At the other end of the spectrum, open-close binary EMG controllers are easier to implement and map directly to one-DOF grippers, but provide less versatility and dexterity. To the best of our knowledge, we are the first to develop an EMG control that can control multiple DOFs for non-anthropomorphic robot hands, combining the dexterity of multi-DOF control with the robustness and cost effectiveness of non-anthropomorphic hands.

\section{Teleoperation Subspace}\label{sec:teleop_subspace}

EMG signals from the forearm are noisy and many of the muscles which control finger movement lie deep under the skin. As a result, it is difficult to project from EMG signals to a high dimensional joint space of a fully actuated robot. It is much more practical to project EMG to a low dimensional subspace relevant to teleoperation, and then project from that subspace to the pose space of a robot hand (Figure~\ref{fig:emg_mapping}). In this work, we use a subspace which we have shown to be relevant for teleoperation~\cite{meeker2018}. We briefly review the mapping between pose space and our teleoperation subspace below.

Teleoperation subspace mapping enables the master and the slave hands to form similar shapes around a scaled object. Teleoperation subspace is a three dimensional subspace $\boldsymbol T$ that can encapsulate the range of movement needed for teleoperation. It is continuous and low dimensional. 

We refer to the three basis vectors of $\boldsymbol T$ as $\boldsymbol\alpha$, $\boldsymbol\sigma$, and $\boldsymbol\epsilon$.  Movement along each of these basis vectors has an intuitive correspondence with hand shape: spreading the fingers ($\boldsymbol\alpha$), scaling the hand to the size of the grasped object ($\boldsymbol\sigma$), and curling the fingers ($\boldsymbol\epsilon$). We have previously shown how to construct mappings from the pose space of the human hand into $\boldsymbol T$ and from $\boldsymbol T$ to the pose space of a robot hand.

In this work, the master's joint angles are unknown and the forearm EMG signals of the teleoperator are known.  To find the joint angles of the slave hand, we must first address how to map forearm EMG signals into $\boldsymbol T$. 

\section{Mapping EMG into Teleoperation Subspace}\label{emg_to_TS}

\begin{figure}[t]
\vspace{0.2cm}
\centering
\begin{tabular}{cc}
\hspace{-6mm}
\includegraphics[trim=3.5cm 13cm 8cm 2.5cm, clip=True,width=1.06\linewidth]{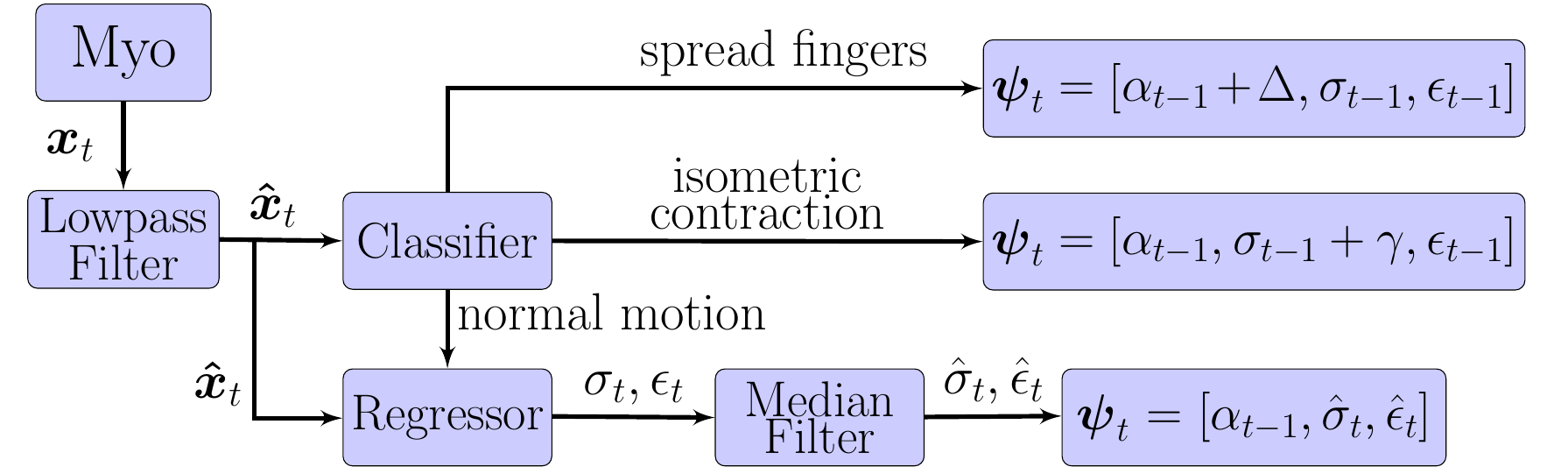}
\vspace{-1mm}
\end{tabular}
\caption{Control scheme for the proposed EMG teleoperation.}
\label{control_scheme}
\vspace{-4mm}
\end{figure}

We map EMG signals into $\boldsymbol T$ using a hybrid method that combines regression (continuous) and classification (discrete). Although we would ideally like to use a regressor to predict the value along all three axes of $\boldsymbol T$, because of the noise inherent in EMG and the distal location of the intrinsic muscles which control finger spread~\cite{schwarz1955}, regression does not always provide an accurate estimation of hand pose in $\boldsymbol T$. We therefore offer the user the option of using two discrete hand signals to move the robot in specific ways. The classifier distinguishes between these two gestures and normal motion of the hand (when the user is not making a gesture). Therefore, classifier has three classes: 
\begin{itemize}
\item Gesture 1 - Finger spread: the user spreads their fingers apart, making the robot fingers spread as well;
\item Gesture 2 - Isometric contraction: the user contracts their muscles and the robot hand closes;
\item Normal motion: the user moves their hand normally, and regression determines the position of the robot. 
\end{itemize}

The regressor is flexible and intuitive, providing a continuous prediction for the size $\boldsymbol\sigma$ and curl $\boldsymbol\epsilon$ bases of $\boldsymbol T$, for which EMG gives a clear signal while the user makes natural grasping motions. The classifier provides stability when the EMG signals are too noisy for regression; it uses discrete signals to control finger spread and to maintain stable grasps, cases where EMG tends to be less reliable. As we will show in our experiments, the combination of continuous and discrete models enables robust and stable teleoperation.

\subsection{Data collection and processing}
In this work, we use the Myo from Thalmic Labs, an EMG armband with 8 EMG sensors. We receive EMG signals $\boldsymbol x_t \in R^8$ from the forearm at time $t$ and use a lowpass filter with a window size of 0.5 seconds to remove noise from the EMG. The filter has a sampling frequency of 5kHz and a cutoff of 200Hz. We refer to the filtered EMG signals as $\boldsymbol{\hat{x}}$.

Our goal at every time step is to find $\boldsymbol \psi_t$, the hand's position in $\boldsymbol T$, given $\boldsymbol{\hat{x}}_t$. To do this, we pass $\boldsymbol{\hat{x}}_t$ through a classifier and then, depending on the output of the classifier, the signals are either passed to a regressor, or the pose from the previous time step is altered in a predefined way. We explain our method below and illustrate it in Figure~\ref{control_scheme}.

\subsection{Classification and Regression}\label{subsec:RFC}

When the classifier identifies that the user is performing one of the two gestures we chose as having meaning for our control method (spreading their fingers, or performing an isometric contraction), our model modifies the pose in $\boldsymbol T$ in a predetermined manner.

When classifier predicts that the teleoperator is spreading their fingers, we freeze the values of $\sigma$ and $\epsilon$ and begin to change the spread value along $\boldsymbol \alpha$. The pose in $\boldsymbol T$ begins with $\alpha$ at its minimum value. When the classifier identifies that the user has spread their fingers, the value of $\alpha$ begins to change at a predetermined rate $\Delta$ until the user stops spreading. $\alpha$ remains at that value until the user spreads their fingers again. $\Delta$ begins as a positive rate, causing the fingers to spread further apart. When the fingers of the robot hand reach their maximum spread value, the sign of $\Delta$ changes, and the fingers begin to move back towards each other. 

The classifier also uses isometric contractions as a discrete signal to help the robot hand close, in order to create more stable grasps. Teleoperators have a natural tendency to perform isometric contractions to ensure the slave robot maintains its grasp. If the model has no special case for this, the muscle contractions can cause a regression model to behave in unexpected ways. When the classifier identifies an isometric contraction, we freeze the values of $\epsilon$ and $\alpha$, the $\sigma$ value output by the regressor is ignored, and $\sigma$ starts to increase at a predetermined rate $\gamma$. The value of $\sigma$ increases until the fingers stall or until the user performs another isometric contraction. If the user performs a second isometric contraction, the regressor resumes predicting the value of $\sigma$ and $\epsilon$, and the classifier resumes controlling the value of $\alpha$.

$\Delta$ and $\gamma$ are set by the experimenter, depending on how fast we want the fingers to spread and the hand to close, respectively.

If the classifier predicts that the teleoperator is moving normally, i.e. not spreading their fingers or performing isometric contractions, the filtered EMG signals $\boldsymbol{\hat{x}}_t$ are passed to a regressor. $\alpha_t$ remains equal to $\alpha_{t-1}$ and the regressor outputs new values for $\sigma_t$ and $\epsilon_t$. 

Our prediction still has some noise after the regression, so, to keep the robot fingers steady, we pass the output of the regressor through a median filter with a window size of 200ms to find our final $\hat\sigma_{t}$ and $\hat\epsilon_{t}$ in $\boldsymbol T$.

\subsection{Robot Pose}

Once we have found the predicted pose in teleoperation subspace $\psi$, we find the joint angles $q$ of the robot as follows: 
\begin{equation}
\boldsymbol q_t = ((\boldsymbol \psi_t \odot \delta^*) \cdot \boldsymbol A^\top) + \boldsymbol o 
\label{eq_from_subspace}
\end{equation}
where $\odot$ represents element-wise multiplication. $\boldsymbol o, \boldsymbol A$ and $\delta^*$ represent an offset, a linear mapping and a scaling factor respectively, all of them specific to the robot hand that is being used. Details on how to build these elements of the mapping can be found elsewhere~\cite{meeker2018}.

\section{Complete Teleoperation Methods} \label{sec:methods}

\begin{figure}[t]
\centering
\vspace{2mm}
\begin{tabular}{r}
\hspace{-4mm}
\includegraphics[trim=5.15cm 8cm 9.1cm 3.47cm, clip, width=1.045\linewidth]{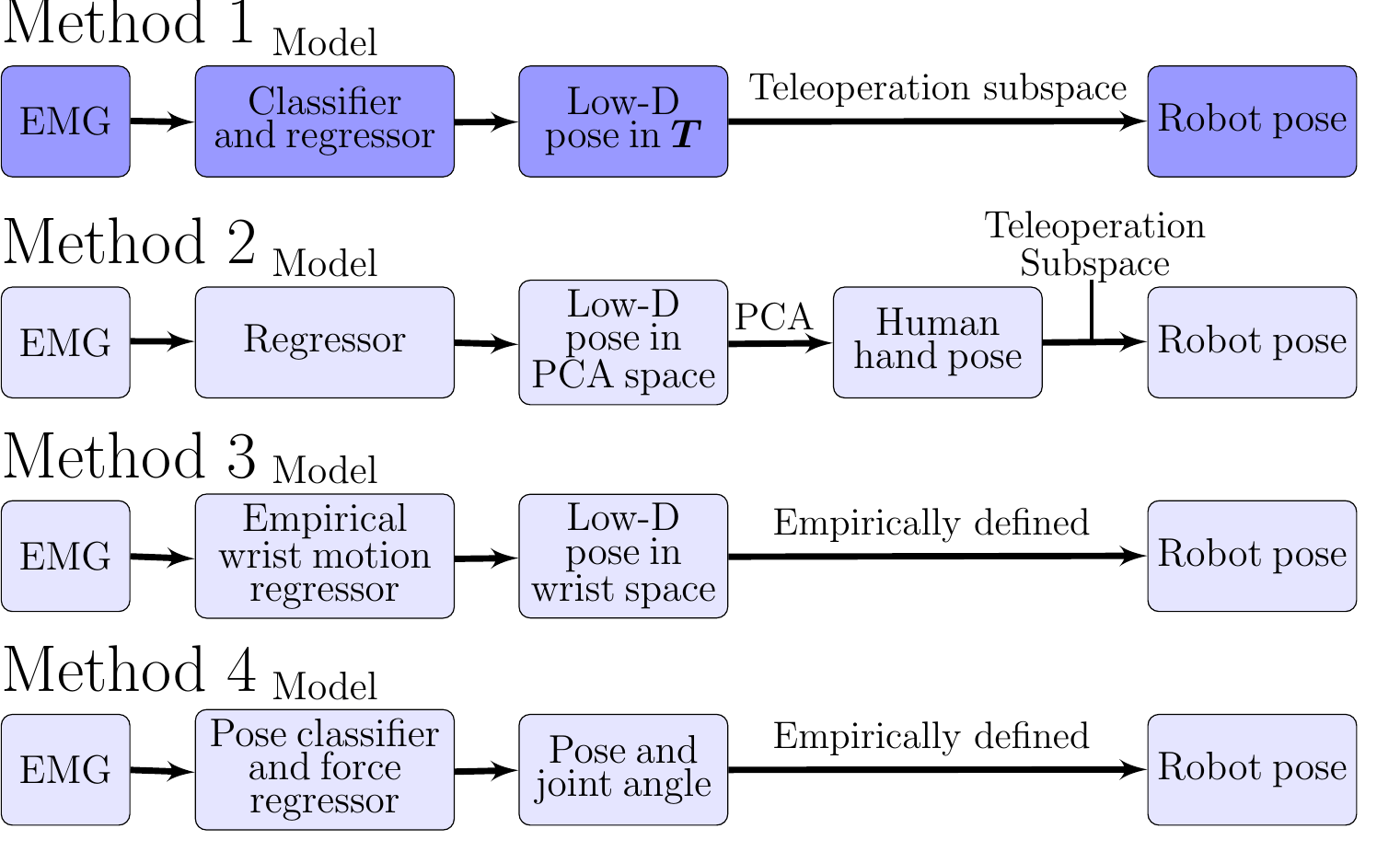}
\end{tabular}
\vspace{-6mm}
\caption{Comparison of control methods which map between forearm EMG and a robot pose space.}
\label{testing_comparison}
\vspace{-6mm}
\end{figure}

We have identified four unique approaches to EMG teleoperation we wish to evaluate experimentally. The first method is our method, which is described above. The next three methods are state-of-the-art EMG teleoperation controls, created for underactuated or anthropomorphic hands. We have modified them as necessary to control non-anthropomorphic, fully actuated hands.  The methods are described below and diagrams of their control structure are shown in Figure~\ref{testing_comparison}.

\subsubsection{Method 1: Regression and Classification using Teleoperation Subspace}\label{train_our_method}
This is our method as described in Section~\ref{emg_to_TS}. It combines a classifier and a regressor to project EMG signals to $\boldsymbol T$ and teleoperation subspace mapping to project from $\boldsymbol T$ to the robot pose space.

\subsubsection{Method 2: Regression using Low-Dimensional Subspace found with PCA} 

For this method, a PCA-based subspace is found by performing PCA on joint angles collected during training. A regressor takes as input filtered forearm EMG signals and outputs a pose in the PCA-based subspace. The predicted poses are projected into human pose space using the PCA components and then mapped into a robotic pose space using teleoperation subspace mapping.

This PCA-based strategy is used in the literature both for underactuated hands~\cite{rossi2017}, and hand/arm systems~\cite{artemiadis2011}. We modified it to work with non-anthropomorphic hands by mapping from the human pose space to the robot pose space with teleoperation subspace mapping.

The basis vectors of the PCA subspace change depending on the training data, so there is no guarantee that the basis vectors will correspond to a given hand motion; therefore, the model cannot combine a classifier and a regressor.

\subsubsection{Method 3: Empirically Defined Regression from EMG to a Low-Dimensional Space using Wrist Motion}\label{sec:method3}

This method uses an empirically defined regressor created by Matrone et al.~\cite{matrone2012}. The regressor projects EMG data into a 2D space. The basis vectors of the space, $C1$ and $C2$, have been shown to correspond well to wrist flexion/extension and wrist abduction/adduction, respectively. The pose in the 2D space is then mapped to robot hand position.

In the original work~\cite{matrone2012}, EMG sensors are placed on two agonistic/antagonistic muscle pairs, one for wrist extension/flexion and one for wrist abduction/adduction. In this work, we place sensors on the same wrist extension/flexion muscle pair, and wrist adduction muscle. However, we are constrained by using an EMG armband, so we cannot place the wrist abduction sensor on the extensor pollicis longus, which is further down the forearm. Instead we place the wrist abduction EMG sensor on the extensor carpi radialis longus.

\subsubsection{Method 4: Pose Classification and Force Regression}

This method, created by Yoshikawa et al.~\cite{yoshikawa2007}, uses a gesture classifier and a force regressor to enable teleoperation.

For the classifier, we selected different hand poses than those presented in the original paper, which mostly classified wrist positions, with only an open and close pose for the hand. We selected three hand poses which represent basic grasp types (power, precision, and pinch). These poses are intuitive because they are distinct from each other and they have a clear corresponding pose in robot pose space.

The force regression is an empirically defined method based on maximum voluntary contraction (MVC) and minimum voluntary contraction for each gesture. The regressor outputs a single joint angle $\theta$, which then must be used to determine hand motion of the robot empirically. We have created our own empirical mapping between $\theta$ and the pose of a non-anthropomorphic robot hand. The greater the isometric contraction, the more the robot hand opens. Low force closes the robot hand because isometric contractions are difficult to maintain over a long period of time.

\section{Experiments}

In this section, we evaluate all of these methods for complete teleoperation with novice users. Since many of them rely on classification or regression, we begin by describing the model training and the selection of model algorithms.

\subsection{Training}

The training process for each of the methods is as follows:

\subsubsection{Method 1} \label{sec:train_method1} 

\begin{table}[]
\vspace{2mm}
\centering
\caption{Normalized root mean square error of trained regressors and global accuracy of trained classifiers.}
\label{fig:model_performance}
\begin{tabular}{cc}
\hspace{-4mm}
\begin{tabular}{C{0.95cm}|C{0.5cm}C{0.5cm}|C{0.6cm}C{0.6cm}||}
 	 & \multicolumn{2}{c|}{Method 1}   & \multicolumn{2}{c||}{Method 2}  \\
Regressor & $\sigma$ 		& $\epsilon$ 		& Basis 1 		 & Basis 2 			\\ \hline &&&&\\[-2mm]
KRR    & 3.7\%   	& \textbf{15.6}\%   & 4.2\%   	 & \textbf{1.9}\% 	\\ 
NMF+LR 	 & 23.4\%   	& 17.3\%   		& 4.8\%   	 & 3.0\%   	\\ 
LS    & \textbf{3.5}\%   & 19.6\%   		& \textbf{3.3}\% & 2.6\%   	\\ 

\end{tabular} &
\hspace{-5.2mm}
\begin{tabular}{C{0.9cm}|C{0.55cm}|C{0.5cm}}
 	 & \multicolumn{2}{c}{Method}   				\\
Classifier & 1 		& 4		\\ \hline &&\\[-2mm]
SVM    & 90.4\%   	& 91.8\% 				\\ 
RF    & \textbf{91.6\%}  & \textbf{94.1\%}   	\\ 

\end{tabular}
\end{tabular}
\vspace{-6mm}
\end{table}

The user generates a training dataset by moving their hand for two minutes while wearing the Myo armband, which provides forearm EMG signals, and a Cyberglove, which provides ground truth joint angles. 

We instructed users to move at a moderate pace and to explore the hand's full range of motion. We also prompted users to perform gestures at 30 second intervals to provide the classifier with training information. 

The control was trained as follows: we passed EMG signals through a bandpass filter and projected joint angles into $\boldsymbol T$ to provide a ground truth values for $\sigma$ and $\epsilon$. We project the joint angles into $\boldsymbol T$ using the equation $\boldsymbol \psi = ((\boldsymbol q-\boldsymbol o) \cdot \boldsymbol A) \odot \delta \label{eq_to_subspace}$~\cite{meeker2018}. We trained the regressor on all data where the user was not performing a gesture. The regressor takes as input the filtered EMG signal and outputs values for $\sigma$ and $\epsilon$. We trained the classifier on all training data. It takes as input the filtered EMG signal and outputs a predicted gesture (or a prediction of no gesture, i.e. normal movement). 

Kernel ridge regressors (KRRs)~\cite{hahne2014}, non-negative factorization (NMF) combined with linear regression (LR)~\cite{rossi2017} and latent space models (LS)~\cite{liarokapis2013} have all been used for EMG controls. We trained three regressors to determine which is best suited to the motions and teleoperation subspace which we use here: using the filtered EMG and the projected values in $\boldsymbol T$, we trained a KRR and a NMF+LS model. For the latent space model, like the original work~\cite{liarokapis2013}, we performed PCA on the EMG signals and projected them into a low-dimensional space. We used the projected EMG data and the poses in $\boldsymbol T$ to train an LS model.

We tested the regressors on three sets of data generated in the same way as the training data. Table~\ref{fig:model_performance}, shows the normalized root mean squared error (nRMSE) of each regressor as a percentage averaged over the three testing datasets.
 
We chose to use a KRR for the rest of our experiments because its nRMSE averaged across $\boldsymbol \sigma$ and $\boldsymbol \epsilon$ was lower than the other two regressors and because KRRs have the ability to update their model with future training data, which can make them robust to drift and donning/doffing, an important issue with EMG controllers~\cite{gijsberts2014}. We do not address model updating here, but we plan to explore this in the future.

The KRR has a radial basis function (RBF) kernel and its alpha and gamma parameters are determined by cross-validation on the data collected to train the model.

We also train two classifiers - a support vector machine (SVM) and a random forest (RF) classifier that can identify the gestures relevant for Method 1 (normal motion, spread, isometric contractions). These have both shown high accuracies when identifying hand gestures~\cite{liarokapis2012}~\cite{yoshikawa2007}. 

The classifiers are trained by the user performing a predefined series of gestures while giving the system ground truth labels for the gesture. We tested each classifier on three testing datasets. Table~\ref{fig:model_performance} shows the average accuracy of both classifiers across the three testing datasets. We chose a RF classifier over a SVM because its accuracy is higher and the time it takes to fit a RF is faster.

\subsubsection{Method 2}
Generating the training dataset for this method is the same process as the training for Method 1, except the user does not perform gestures intermittently. 

Once the dataset is gathered, we perform PCA on the joint angles from the Cyberglove and keep the eigenvectors which explain 90\% of the variance in the data. The joint angles are projected to the PCA-based space with the eigenvectors. 

The EMG is put through the same bandpass filter used in Method 1. The model for this method takes filtered EMG as an input and outputs a pose in the PCA-based space.

As for Method 1, we wished to determine which of the three regression models presented in the literature work best with a PCA-based subspace. We trained KRR, NMF+LR and LS models in the same way as for Method 1, but using low-dimensional poses in the PCA subspace as ground truth. 

The Method 2 regressors were also tested on three datasets (Table~\ref{fig:model_performance}). We again chose to use the KRR because of the possibility of updating the model with future data.

\subsubsection{Method 3}

Method 3 uses an empirically defined regression which maps EMG signals to a two dimensional space whose basis vectors have been shown to correspond well with certain wrist motions. Because the regression is empirically defined, the training only requires four calibration poses - flexing, extending, abducting and adducting the wrist to provide the MVC for each of the gestures.

In the original work that outlined this method, medical grade sensors were placed by trained experimenters on specific muscle pairs. Our EMG armband is inexpensive and easy to don even by a novice, but also provides constraints: lower quality signal, and the need to use muscles that are all at similar height on the forearm. Within these constraints, our experiments showed that the basis vector $C2$ no longer corresponds to the expected wrist motion (abduction/adduction). 

We asked a user to perform three different wrist gestures and projected the collected EMG data into the 2D subspace. To see if the basis vectors corresponded to the expected wrist motions, we compared the variance along the basis vectors. The variance along $C1$ is 0.0 when the user holds their wrist still and increases 66-fold when the user performs wrist flexion/extension. On the other hand, the variance along $C2$ is 0.0 when the user holds their wrist still, 0.02 when the user abducts/adducts, and 0.02 when the user flexes/extends. We conclude that, within the constraints of a commercial EMG armband, this method is ineffective because $C2$ does not correspond to the expected wrist motion.

\subsubsection{Method 4}

Method 4 uses a classifier to determine hand pose and an empirically defined force regressor to determine a joint angle $\theta$. The predicted hand pose and $\theta$ map to robot hand position. We trained the classifier by asking the user to perform a series of gestures. During these gestures, the user performed several isometric contractions and then relaxed. 

The user also had to perform six calibration gestures to train the empirically defined force regressor. They performed each of the three hand poses while they were relaxed, and while they were performing isometric contractions.

As for Method 1, we wished to determine if an SVM or an RF classifier would better distinguish between the classes relevant for Method 4 (pinch, spread, parallel grasp). We trained two more classifiers and tested in the same way as for Method 1. The average accuracy of the two classifiers are shown in Table~\ref{fig:model_performance}. We again chose to use a RF classifier because of its higher accuracy and shorter fit time.

\begin{figure*}[t]
\centering
\vspace{2mm}
\hspace{-8mm}
\begin{tabular}{C{2.0cm}C{3.3cm}C{12.4cm}}
\hspace{-6mm}
\includegraphics[trim=2.5cm 0.6cm 2.5cm 4cm, clip,width=1\linewidth]{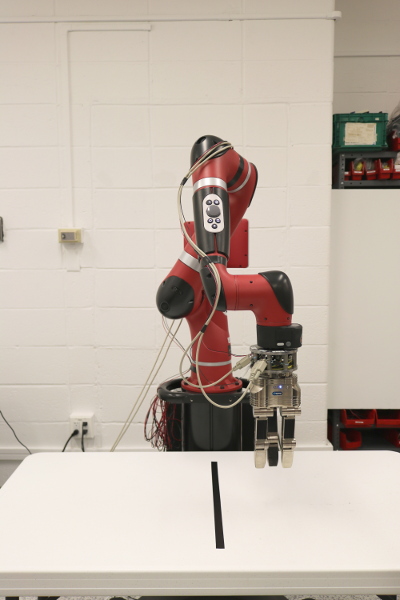} &
\hspace{-10mm}
\includegraphics[trim=0cm 0cm 0cm 0cm, clip, width=1.06\linewidth]{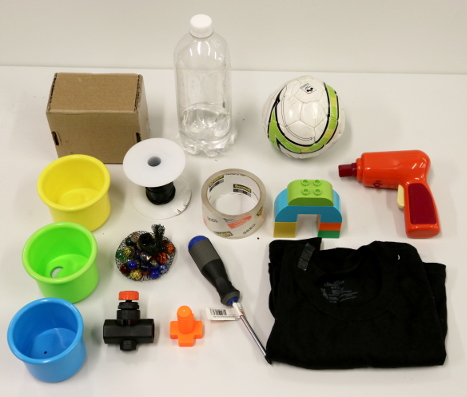} &
\hspace{-16mm}
\includegraphics[trim=8.6cm 7.65cm 0cm 4.65cm, clip, width=1\linewidth]{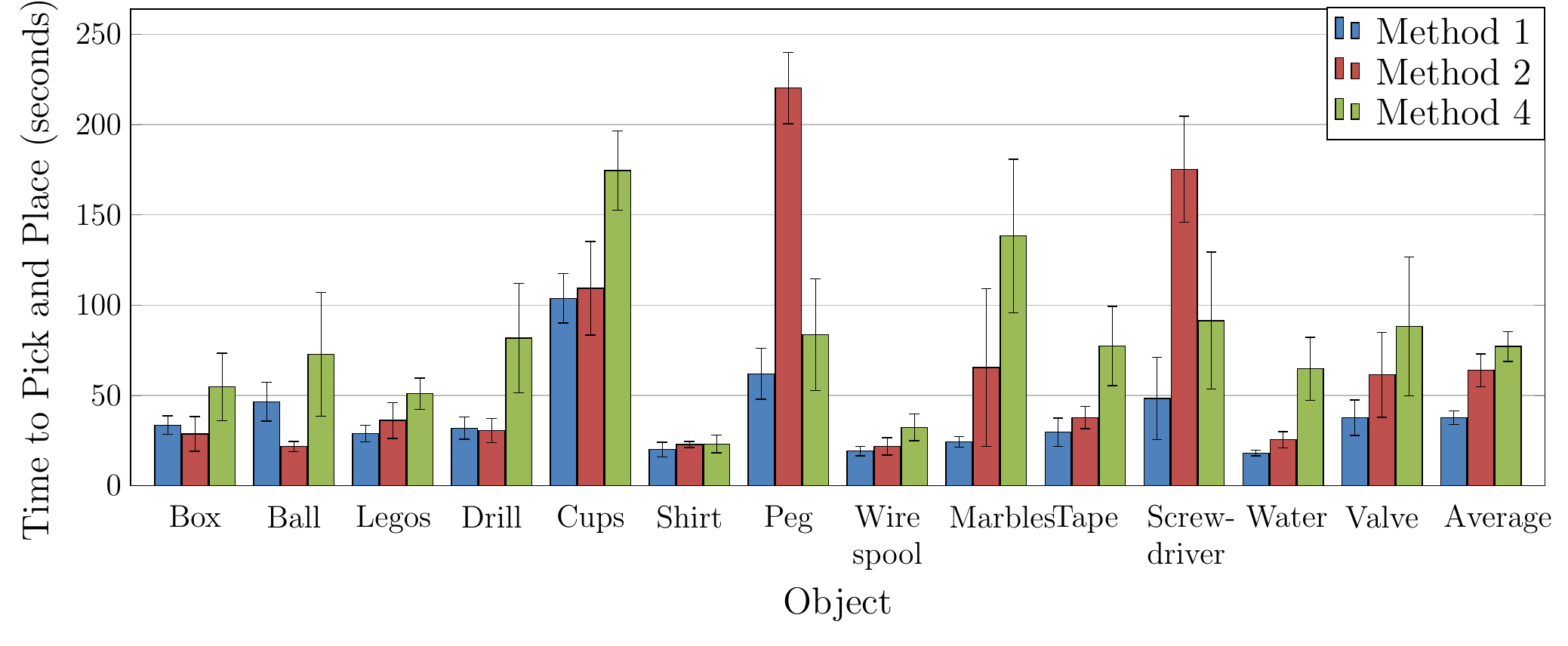}
\end{tabular}
\vspace{-6mm}
\caption{(Left) Experimental set-up, (middle) object set and (right) time to pick and place each object for pick and place experiments.}
\label{fig:pick_and_place_tests}
\vspace{-5mm}
\end{figure*}

\subsection{Complete Pick and Place with Novice Users}\label{hybrid_experiments}

To show that EMG controls enable effective teleoperation of non-anthropomorphic hands, we asked five novice to pick and place a variety of objects using Methods 1, 2 and 4 above (we excluded Method 3 since, as discussed above, it proved not applicable with a commercial EMG armband). The subjects gave their informed consent and the study was approved by the Columbia University IRB.

The hand used for teleoperation was the Schunk SDH hand attached to a Sawyer Robot arm (Figure~\ref{fig:pick_and_place_tests}). The Sawyer's end effector position and orientation are controlled with a simple cartesian controller (completely separate from the hand control) using a magnetic tracker (Ascension 3D Guidance trakSTAR\texttrademark) placed on the back of the user's hand. Users teleoperated based on visual feedback.

The users were asked to complete 13 pick and place tasks (the object set is shown in Figure~\ref{fig:pick_and_place_tests}). The user was either asked to pick up the item and move it across a line or, in the case of the three cups, stack the items. The user completed the pick and places for all objects with one control method before moving to the next control. The order in which each subject used the control methods was randomized. 

We timed how long it took for the users complete each pick and place. If a user did not complete the task in four minutes, they were considered to be unable to pick up the object and their final time was set to four minutes.

\subsection{Results}

Across all subjects and all objects, with Method 2, it took novices 1.70 times longer to complete a pick and place than when they were using Method 1. Method 4 took 2.05 times longer than Method 1. Figure~\ref{fig:pick_and_place_tests} shows the average time to pick and place and Table~\ref{tab:pick_and_place_stats} summarizes the average statistics for each method calculated across all subjects and all objects.

Let us consider the pick and place times more granularly. We hypothesized that Method 1 would help most for small, circular objects. If we only consider the valve, the marbles, the orange peg, and the screwdriver, the average time to pick and place using Method 2 was 3.43 times longer than Method 1. With Method 4, it was 2.33 times longer than with Method 1. If we consider all objects but the four mentioned above (the larger objects), the average time to pick and place with Method 2 was 1.07 times longer than Method 1 and with Method 4 was 1.95 times longer than Method 1.

Method 1 was the only control method which allowed all the novice users to pick up all 13 objects. With Method 2, the subjects were able to pick up, on average, 11.6 objects, and with Method 4, subjects averaged 12.2 objects. 

We counted the number of tries it took for the user to pick up each object. We define a `try' as a completed pick and place task, an attempt where the user drops the object or an attempt where the user knocks over the object. With Method 1, the average number of tries across all subjects and all objects was 1.2. With Method 2, the average number of tries was 1.8 and with Method 4, it was 1.6.

\begin{table}[t]
\vspace{2mm}
\centering
\caption{Pick and place statistics}
\label{tab:pick_and_place_stats}
\vspace{-2mm}
\begin{tabular}{C{0.7cm}|C{1.7cm}|C{1.4cm}|C{1.8cm}}
Method  & Time to completion (s)  	& Tries per object				& Successful picks (max 13)		\\ \hline &&&\\[-2mm]
1  	& \textbf{37.6 $\pm$ 3.8}		& \textbf{1.2 $\pm$ 0.1}	&\textbf{13 $\pm$ 0.0}			   \\ 
2  	& 63.9 $\pm$ 9.1    			& 1.8 $\pm$ 0.2      		&11.6 $\pm$ 0.5				 \\ 
4  	& 77.1 $\pm$ 8.3     			& 1.6 $\pm$ 0.2      		&12.2 $\pm$ 0.6				 \\ 

\end{tabular}
\vspace{-6mm}
\end{table}

\section{Discussion}

Our experiments show that EMG teleoperation is possible with a fully actuated, non-anthropomorphic hand. When compared with two state-of-the-art EMG teleoperation methods, novices were able to use Method 1 (ours) to pick up a wider variety of objects, faster and with fewer mistakes, than they could with Methods 2 and 4.

Our method provides the greatest advantage when attempting to pick and place smaller objects. With larger objects, Method 1 and Method 2 enable teleoperation at about the same speed. We hypothesize that our hybrid method provides this advantage because it is difficult for the user to both close their fingers and also spread them. Our hybrid model provides explicit control over the spread and close motions of the robot hand, thus avoiding awkward hand positions and making stable grasping easier. Our hybrid method works with the teleoperation subspace because the consistent basis vectors allow the use of both discrete and continuous models. 

We notice in Figure~\ref{fig:pick_and_place_tests} that, for some of the objects, the standard error is high, particularly when the average time to pick and place was over 50 s. For the smaller objects, this is usually because some subjects were able to pick up the object while others were not able to do so in the allotted four minutes. For larger objects, the high error is usually associated with Method 4. Some subjects found it difficult to create hand poses which were easily distinguishable for the classifier and therefore took much longer to pick and place than subjects for whom the pose classifier worked well. 

We have previously reported the average time to pick and place for novice users with teleoperation subspace mapping controlled by a dataglove was 27.5 s and the average time to pick and place with other state-of-the-art teleoperation mappings was between 62.3 and 56.7 s~\cite{meeker2018}. When compared to EMG teleoperation, we see that Method 1 enables teleoperation faster than the state-of-the-art methods controlled by a dataglove, and only 10 s slower than teleoperation subspace mapping controlled by a dataglove. Methods 2 and 4 enable teleoperation at about the same speed as state-of-the-art dataglove teleoperation, but much slower than subspace mapping teleoperation. As expected, EMG control is slower than our dataglove control, as the user has to perform additional gestures (spread and isometric contractions) during grasping. The fact that the EMG control is still faster than state-of-the-art methods controlled by a dataglove demonstrates the usefulness of the teleoperation subspace. 

\section{Conclusions and Future Work}

We introduced here a method for teleoperating a non-anthropomorphic, fully actuated robot with forearm EMG. We use a subspace relevant to teleoperation as an intermediary between EMG and robot pose space, and combine continuous and discrete models for control through this subspace. To the best of our knowledge, we show the first instance of EMG controlled teleoperation of a non-anthropomorphic, fully actuated robot hand.

We compared our control to other state-of-the-art EMG teleoperation methods, which we modified as needed to work with non-anthropomorphic hands. Our method allows users to form stable grasps around a variety of objects faster and with fewer tries than the state-of-the-art methods. We also enable teleoperation faster than state-of-the-art teleoperation controlled with datagloves, and only slightly slower than dataglove-controlled teleoperation subspace mapping.

Although the control we describe here requires a dataglove to train the EMG regressor, the dataglove is removed when our teleoperation method is used in practice. EMG armbands are less expensive and easier to replace than datagloves. Furthermore, their placement on the forearm, rather than on the hand, makes them less susceptible to damage and leaves the hand completely unencumbered as the teleoperator performs collaborative tasks with the robot. 

In the future, we would like to make our method more robust to donning and doffing by updating our EMG models with additional training data. We would also like to perform more experiments testing novice users' ability to perform more complex manipulation with EMG teleoperation.

\bibliographystyle{IEEEtran}
\bibliography{bib/teleoperation}

\begin{thebibliography}{10}
\providecommand{\url}[1]{#1}
\csname url@rmstyle\endcsname
\providecommand{\newblock}{\relax}
\providecommand{\bibinfo}[2]{#2}
\providecommand\BIBentrySTDinterwordspacing{\spaceskip=0pt\relax}
\providecommand\BIBentryALTinterwordstretchfactor{4}
\providecommand\BIBentryALTinterwordspacing{\spaceskip=\fontdimen2\font plus
\BIBentryALTinterwordstretchfactor\fontdimen3\font minus
  \fontdimen4\font\relax}
\providecommand\BIBforeignlanguage[2]{{%
\expandafter\ifx\csname l@#1\endcsname\relax
\typeout{** WARNING: IEEEtran.bst: No hyphenation pattern has been}%
\typeout{** loaded for the language `#1'. Using the pattern for}%
\typeout{** the default language instead.}%
\else
\language=\csname l@#1\endcsname
\fi
#2}}

\bibitem{ferre2007}
M.~Ferre, R.~Aracil, C.~Balaguer, M.~Buss, and C.~Melchiorri, \emph{Advances in
  telerobotics}.\hskip 1em plus 0.5em minus 0.4em\relax Springer, 2007,
  vol.~31.

\bibitem{rohling1993}
R.~N. Rohling, J.~M. Hollerbach, and S.~C. Jacobsen, ``Optimized fingertip
  mapping: a general algorithm for robotic hand teleoperation,''
  \emph{Presence: Teleoperators \& Virtual Environments}, vol.~2, no.~3, pp.
  203--220, 1993.

\bibitem{cerulo2017}
I.~Cerulo, F.~Ficuciello, V.~Lippiello, and B.~Siciliano, ``Teleoperation of
  the schunk s5fh under-actuated anthropomorphic hand using human hand motion
  tracking,'' \emph{Robotics and Autonomous Systems}, vol.~89, pp. 75--84,
  2017.

\bibitem{ekvall2004}
S.~Ekvall and D.~Kragic, ``Interactive grasp learning based on human
  demonstration,'' in \emph{Robotics and Automation (ICRA), 2004 IEEE Int Conf
  on}, vol.~4.\hskip 1em plus 0.5em minus 0.4em\relax IEEE, 2004, pp.
  3519--3524.

\bibitem{meeker2018}
C.~Meeker, T.~Rasmussen, and M.~Ciocarlie, ``Intuitive hand teleoperation by
  novice operators using a continuous teleoperation subspace,'' in
  \emph{Robotics and Automation (ICRA), 2018 IEEE Int Conf on}, 2018.

\bibitem{herrera2004}
A.~Herrera, A.~Bernal, D.~Isaza, and M.~Adjouadi, ``Design of an electrical
  prosthetic gripper using emg and linear motion approach,'' in
  \emph{Proceedings from the 17th Florida Conference on the Recent Advances in
  Robotics (FCRAR), Florida, USA}, 2004.

\bibitem{gillespie2010}
R.~B. Gillespie, J.~L. Contreras-Vidal, P.~A. Shewokis, M.~K. O'Malley, J.~D.
  Brown, H.~Agashe, R.~Gentili, and A.~Davis, ``Toward improved sensorimotor
  integration and learning using upper-limb prosthetic devices,'' in
  \emph{Engineering in Medicine and Biology Society (EMBC), 2010 Annual Int
  Conf on}.\hskip 1em plus 0.5em minus 0.4em\relax IEEE, 2010, pp. 5077--5080.

\bibitem{vasan2017}
G.~Vasan and P.~M. Pilarski, ``Learning from demonstration: Teaching a
  myoelectric prosthesis with an intact limb via reinforcement learning,'' in
  \emph{Rehabilitation Robotics (ICORR), 2017 Int Conf on.}\hskip 1em plus
  0.5em minus 0.4em\relax IEEE, 2017, pp. 1457--1464.

\bibitem{choudhary2012}
S.~K. Choudhary, D.~Chakraborty, N.~M. Kakoty, and S.~M. Hazarika,
  ``Development of cost effective emg controlled three fingered robotic hand,''
  in \emph{Computer and Communication Technology (ICCCT), 2012 Third
  International Conference on}.\hskip 1em plus 0.5em minus 0.4em\relax IEEE,
  2012, pp. 104--109.

\bibitem{fani2016}
S.~Fani, M.~Bianchi, S.~Jain, J.~S. Pimenta~Neto, S.~Boege, G.~Grioli,
  A.~Bicchi, and M.~Santello, ``Assessment of myoelectric controller
  performance and kinematic behavior of a novel soft synergy-inspired robotic
  hand for prosthetic applications,'' \emph{Frontiers in neurorobotics},
  vol.~10, p.~11, 2016.

\bibitem{malesevic2017}
N.~Male{\v{s}}evi{\'c}, D.~Markovi{\'c}, G.~Kanitz, M.~Controzzi, C.~Cipriani,
  and C.~Antfolk, ``Decoding of individual finger movements from surface emg
  signals using vector autoregressive hierarchical hidden markov models
  (varhhmm),'' in \emph{Rehabilitation Robotics (ICORR), 2017 Int Conf on.},
  2017, pp. 1518--1523.

\bibitem{smith2008}
R.~J. Smith, F.~Tenore, D.~Huberdeau, R.~Etienne-Cummings, and N.~V. Thakor,
  ``Continuous decoding of finger position from surface emg signals for the
  control of powered prostheses,'' in \emph{Engineering in Medicine and Biology
  Society (EMBC), 2008 Annual Int Conf on}.\hskip 1em plus 0.5em minus
  0.4em\relax IEEE, 2008, pp. 197--200.

\bibitem{hahne2014}
J.~M. Hahne, F.~Biessmann, N.~Jiang, H.~Rehbaum, D.~Farina, F.~Meinecke, K.-R.
  M{\"u}ller, and L.~Parra, ``Linear and nonlinear regression techniques for
  simultaneous and proportional myoelectric control,'' \emph{IEEE Trans Neural
  Syst Rehabil Eng}, vol.~22, pp. 269--279, 2014.

\bibitem{jiang2009}
N.~Jiang, K.~B. Englehart, and P.~A. Parker, ``Extracting simultaneous and
  proportional neural control information for multiple-dof prostheses from the
  surface electromyographic signal,'' \emph{IEEE Trans Biomed Eng}, vol.~56,
  no.~4, pp. 1070--1080, 2009.

\bibitem{jiang2014}
N.~Jiang, H.~Rehbaum, I.~Vujaklija, B.~Graimann, and D.~Farina, ``Intuitive,
  online, simultaneous, and proportional myoelectric control over two
  degrees-of-freedom in upper limb amputees,'' \emph{IEEE Trans Neural Syst
  Rehabil Eng}, vol.~22, no.~3, pp. 501--510, 2014.

\bibitem{lin2017}
C.~Lin, B.~Wang, N.~Jiang, and D.~Farina, ``Robust extraction of basis
  functions for simultaneous and proportional myoelectric control via sparse
  non-negative matrix factorization,'' \emph{J Neural Eng}, 2017.

\bibitem{yoshikawa2007}
M.~Yoshikawa, M.~Mikawa, and K.~Tanaka, ``Hand pose estimation using emg
  signals,'' in \emph{Engineering in Medicine and Biology Society (EMBC), 2007
  Annual Int Conf on}.\hskip 1em plus 0.5em minus 0.4em\relax IEEE, 2007, pp.
  4830--4833.

\bibitem{yamanoi2017}
Y.~Yamanoi, S.~Morishita, R.~Kato, and H.~Yokoi, ``Development of myoelectric
  hand that determines hand posture and estimates grip force simultaneously,''
  \emph{Biomedical Signal Processing and Control}, vol.~38, pp. 312--321, 2017.

\bibitem{castellini2009}
C.~Castellini and P.~van~der Smagt, ``Surface emg in advanced hand
  prosthetics,'' \emph{Biological cybernetics}, vol. 100, no.~1, pp. 35--47,
  2009.

\bibitem{gijsberts2014}
A.~Gijsberts, R.~Bohra, D.~Sierra~Gonz{\'a}lez, A.~Werner, M.~Nowak, B.~Caputo,
  M.~A. Roa, and C.~Castellini, ``Stable myoelectric control of a hand
  prosthesis using non-linear incremental learning,'' \emph{Frontiers in
  neurorobotics}, vol.~8, p.~8, 2014.

\bibitem{santello2016}
M.~Santello, M.~Bianchi, M.~Gabiccini, E.~Ricciardi, G.~Salvietti,
  D.~Prattichizzo, M.~Ernst, A.~Moscatelli, H.~J{\"o}rntell, A.~M. Kappers,
  \emph{et~al.}, ``Hand synergies: integration of robotics and neuroscience for
  understanding the control of biological and artificial hands,'' \emph{Physics
  of life reviews}, vol.~17, pp. 1--23, 2016.

\bibitem{rossi2017}
M.~Rossi, C.~Della~Santina, C.~Piazza, G.~Grioli, M.~Catalano, and A.~Biechi,
  ``Preliminary results toward a naturally controlled multi-synergistic
  prosthetic hand,'' in \emph{Rehabilitation Robotics (ICORR), 2017 Int Conf
  on.}\hskip 1em plus 0.5em minus 0.4em\relax IEEE, 2017, pp. 1356--1363.

\bibitem{matrone2012}
G.~C. Matrone, C.~Cipriani, M.~C. Carrozza, and G.~Magenes, ``Real-time
  myoelectric control of a multi-fingered hand prosthesis using principal
  components analysis,'' \emph{J Neuroeng Rehabil}, vol.~9, no.~1, p.~40, 2012.

\bibitem{artemiadis2011}
P.~K. Artemiadis and K.~J. Kyriakopoulos, ``A switching regime model for the
  emg-based control of a robot arm,'' \emph{IEEE Trans Syst Man Cybern B
  Cybern}, vol.~41, no.~1, pp. 53--63, 2011.

\bibitem{liarokapis2013}
M.~V. Liarokapis, P.~K. Artemiadis, K.~J. Kyriakopoulos, and E.~S. Manolakos,
  ``A learning scheme for reach to grasp movements: on emg-based interfaces
  using task specific motion decoding models,'' \emph{IEEE J Biomed Health
  Inform}, vol.~17, no.~5, pp. 915--921, 2013.

\bibitem{schwarz1955}
R.~J. Schwarz and C.~Taylor, ``The anatomy and mechanics of the human hand,''
  \emph{Artificial limbs}, vol.~2, no.~2, pp. 22--35, 1955.

\bibitem{liarokapis2012}
M.~V. Liarokapis, P.~K. Artemiadis, P.~T. Katsiaris, K.~J. Kyriakopoulos, and
  E.~S. Manolakos, ``Learning human reach-to-grasp strategies: Towards
  emg-based control of robotic arm-hand systems,'' in \emph{Robotics and
  Automation (ICRA), 2012 IEEE Int Conf on}, 2012, pp. 2287--2292.

\end{thebibliography}

\end{document}